\documentclass[wcp]{jmlr}
\usepackage{longtable}
\usepackage{amssymb}
\usepackage{graphicx}
\usepackage{amsmath}
\usepackage{hyperref}

\usepackage{url}
\urldef{\mailsa}\path|{phelahab@gmail.com}|
\urldef{\mailsb}\path|anna.kramer, leonie.kunz, christine.reiss, nicole.sator,|

\usepackage{algorithm}

\makeatother

\begin{document}

\jmlrvolume{80}
\jmlryear{2017}
\jmlrworkshop{ACML 2017}

    \title[Neural Stain-Style Transfer Network]{Neural Stain-Style Transfer Learning using GAN for Histopathological Images}

    \author{\Name{Hyungjoo Cho}\thanks{Equal contribution} \Email{phelahab@gmail.com}\\
    \addr Seoul National University
    \AND
    \Name{Sungbin Lim$^{*}$} \Email{sungbin@korea.ac.kr}\\
    \addr Korea University
    \AND
    \Name{Gunho Choi} \Email{ghc0311@gmail.com}\\
    \addr Yonsei University
    \AND
    \Name{Hyunseok Min} \Email{min6284@gmail.com}\\
    \addr KAIST
     }
        
    
    
    %

    \maketitle
    \begin{abstract}

        
        Performance of data-driven network for tumor classification varies with stain-style of histopathological images. This article proposes the \emph{stain-style transfer} (SST) model based on conditional generative adversarial networks (GANs) which is to learn not only the certain color distribution but also the corresponding histopathological pattern. Our model considers \emph{feature-preserving} loss in addition to well-known GAN loss. Consequently our model does not only transfers initial stain-styles to the desired one but also prevent the degradation of tumor classifier on transferred images. The model is examined using the CAMELYON16 dataset.

        
    \end{abstract}

    
    \section{Introduction}

    
    %
    
    Deep learning based image recognition receives a lot of attention due to its notable application to digital histopathology including automatic tumor classification. Convolutional neural networks(CNNs) have recently achieved state-of-the-art performance in the task of image classification and detection, especially, replaced the traditional rule-based methods in the several contests of medical image diagnosis \cite{lecun2015deep, wang2016deep}. Such data-driven approach especially depends on quality of training dataset hence it requires sensible preprocesses. In histopathology, staining e.g. haematoxylin and eosin (H\&E) is essential to examine the microscopic presence and characteristics of disease not only for pathologists but also for neural networks. For digital histopathology, several stain normalization preprocesses are well-known \cite{ruifrok2001quantification, reinhard2001color, ruifrok2003comparison, annadurai2007fundamentals, magee2009colour, macenko2009method, khan2014nonlinear, li2015complete, bejnordi2016stain}. 
     
          \begin{figure}[t]
        \centering
        \includegraphics[width=1\textwidth]{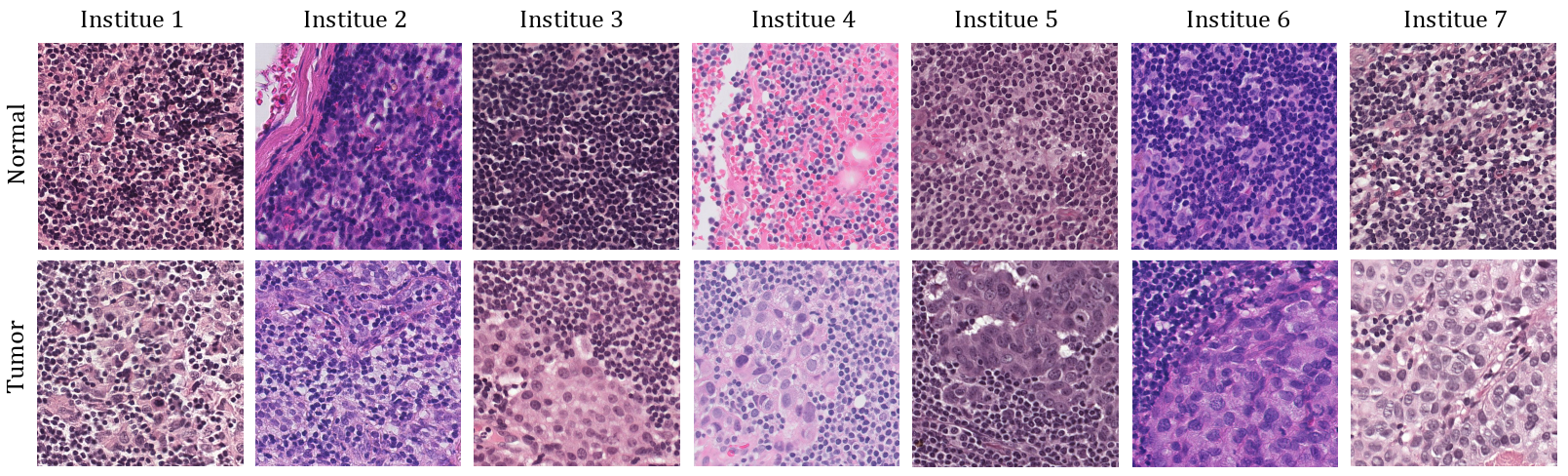}
        \caption{Samples of tissue tiles from different institutes in CAMELYON16\cite{cam16}, 17\cite{cam17} dataset. The first row shows normal samples, and the second row shows tumor samples. Samples of institute 1, 2 are included \cite{cam16}, the others are included \cite{cam17} dataset.}
        \label{fig:=-=}
    \end{figure}
     
     Standard stain normalization algorithms are based on stain-specific color deconvolution \cite{ruifrok2001quantification}. Stain deconvolution requires prior knowledge of reference stain vectors for every dye present in the whole-slide images (WSI). \cite{ruifrok2001quantification} suggested a manual approach to estimate the color deconvolution vectors by selecting representative sample pixels from each stain class, and a similar approach was used in \cite{magee2009colour} for extracting stain vectors. Such manual estimation of stain vectors, however, strongly limits their applicability in large studies. \cite{khan2014nonlinear} modified \cite{magee2009colour} by estimating stable stain matrices using an image-specific color descriptor. Combined with a robust color classification framework based on a variety of training data from a particular stain with nonlinear channel mappings, the method ensured smooth color transformation without introducing visual artifacts. Another approach used in \cite{bejnordi2016stain} transforms the chromatic and density distributions for each individual stain class in the hue-saturation-density (HSD) color model. See \cite{bejnordi2016stain} and references therein. 

     Stain normalization methods for histopathological images have been studied extensively, and yet these still possess challenging issues. Most of the conventional methods use various thresholds to filter out backgrounds and other irrelevant dimensions. However, these methods cannot represent the broad feature distribution of the entire target image, thus they require manual tuning of hyper-parameters such as thresholds. Furthermore, since nuclei detection has a significant impact on performance of color normalization, it is unlikely to expect good performance if there is a mistake in the nuclei detection stage. Finally, although the major aim of most conventional approaches is to enhance the prediction performance of classification system, these stain normalization methods and classifer work separately. It is reported that performance of network varies with institutes even they applied same staining methods \cite{ciompi2017importance}. In order to prevent such variation, it is required to consider a domain adaption method. 
     
     In this paper, we propose a novel stain-style transfer method using deep learning, as well as a special loss function which minimizes the difference between latent features of input image and that of target image, thus preserves the performance of the classifier. We implement fully convolutional network  (FCNs) \cite{long2015fully} in proposed stain-style generator that learns the color distribution of dataset which is used to train the tumor classifier.
     
     Our contributions in this paper are of two areas. First, we replace the color normalization methods with a generative model which learns certain stain-style distribution of dataset. Second, we introduce feature-preserving loss to induce the classifier to extract better features than different methods.

    \section{Stain-Style Transfer with GAN}

    \subsection{Stain-Style of Dataset}

    In this section, we summarize relevant material on our model. Let $I$ be a set of institutes and let $x \in \mathcal{X}$ be the dataset of histological sample and the corresponding label $y\in\mathcal{Y}$. The class of stained images or color images with RGB channels, denoted by $\mathcal{C} = \mathcal{M}_{d}(\mathbb{R}^{3})$, is defined to be the set of $d\times d$-matrix with $\mathbb{R}^{3}$ entries. Under this setting, we define the \emph{stain-style} of institute $i\in I$ to be a random variable $\phi^{(i)} : \mathcal{X} \to \mathcal{C}$ with a probability distribution 
    
    $$
    \mathbb{P}^{(i)}(c):=\mathbb{P}(\phi^{(i)}= c),\quad c\in\mathcal{C}
    $$
    Since $\phi^{(i)}$ admits a certain conditional probability distribution $\mathbb{P}^{(i)}(\cdot|y)$, the definition of different stain-style with the same label makes sense. 
    
    \subsection{Tumor Classifier Network}

    Suppose we trained \emph{tumor classifier network} $f:\mathcal{X} \to \mathcal{Y} $ which infers histological pattern of input image $x\in \mathcal{X}$. We write $f = f^{(i)}$ if the classifier is especially trained on dataset which follows stain-style $\phi^{(i)}$. We estimate the performance of $f^{(i)}$ by 
    \begin{align}\label{original}
    \mathcal{L}^{(i)}_{\text{tumor}}:=\mathbb{E}_{(x,y)\sim P(x,y)}\left[ \ell(f^{(i)}(x),y) \right]
    \end{align}
    where $\ell$ is a loss function for classification e.g. cross-entropy. Practically, we make classifier to learn stained images $\phi^{(i)}(x) \in \mathcal{C}$ rather than $x\in\mathcal{X}$. Hence one can decompose the classifier by $f^{(i)} = \hat{f}^{(i)}\circ\phi^{(i)}$ where $\hat{f}^{(i)} : \mathcal{C}\to\mathcal{Y}$ is an actual network which is trained on dataset with stain-style $\phi^{(i)}$. In this case, we estimate \eqref{original} by
    
    \begin{align}\label{practical}
    \mathcal{L}^{(i)}_{\text{tumor}}\approx\mathbb{E}_{c\sim\mathbb{P}^{(i)}(c|y)}\left[ \ell(\hat{f}^{(i)}(c),y) \right]
    \end{align}
    
    \subsection{Stain-Style Transfer Network}

    \begin{figure}[t]
        \centering
        \includegraphics[width=1\textwidth]{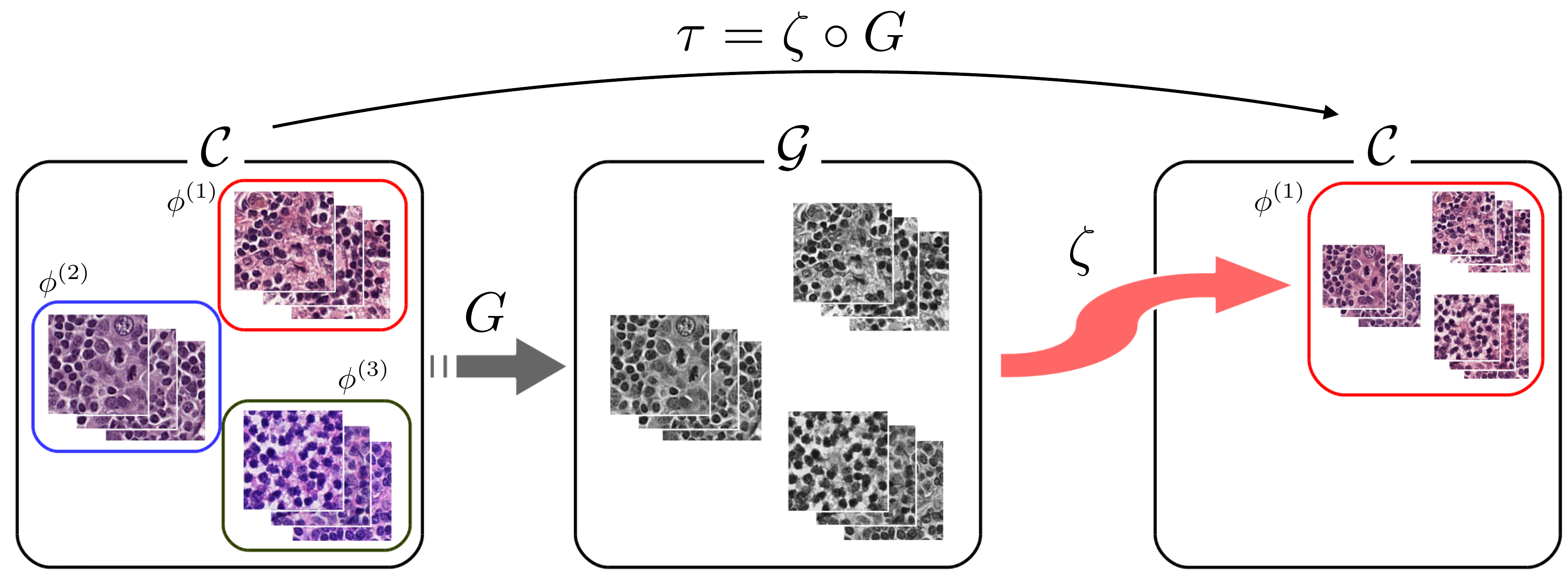}
        \caption{Overview of the stain-style transfer network. The network $\tau$ is composed of two transformations: Gray-normalization $G$ and style-generator $\zeta$. $G$ standardizes each stain-style of color images from different institutes and $\zeta$ colorizes gray images following the stain-style of certain institute.}
        \label{fig:overview}
    \end{figure}

    Since the stain-styles of each institute are dissimilar, the histological pattern in image from different institute would break up in the view of classifier network. Consequently, it would show degraded performance \cite{ciompi2017importance}:
    $$
    \mathbb{E}_{c\sim\mathbb{P}^{(j)}(c|y)}\left[ \ell(\hat{f}^{(i)}(c),y) \right] \geq \mathbb{E}_{c\sim\mathbb{P}^{(i)}(c|y)}\left[ \ell(\hat{f}^{(i)}(c),y) \right],\quad i\neq j
    $$
    To overcome this problem, we propose \emph{stain-style transfer} (SST) network which transfers stain-style $\phi^{(j)}$ to the initial $\phi^{(i)}$. Precisely, our aim is to find a network $\tau : \mathcal{C} \to \mathcal{C}$ which satisfies 
    \begin{align}\label{transfer}
    \mathbb{P}(\tau\circ\phi^{(j)}=c|y)=\mathbb{P}^{(j)}(\tau^{-1}(c)|y)\approx\mathbb{P}^{(i)}(\cdot|y),\quad \forall c\in\mathcal{C}
    \end{align}
    Due to the change of variable formula \cite[Theorem 1.6.9]{durrett2010probability}, \eqref{transfer} implies 
    \begin{align}\label{approx}
    \mathbb{E}_{c\sim\mathbb{P}^{(j)}(\tau^{-1}(\cdot)|y)}\left[ \ell(\hat{f}^{(i)}(c),y) \right] &\approx\mathbb{E}_{\tau(c)\sim\mathbb{P}^{(i)}(\cdot|y)}\left[ \ell(\hat{f}^{(i)}(\tau(c)),y) \right]
    \\ &= \mathbb{E}_{c\sim\mathbb{P}^{(i)}(\cdot|y)}\left[ \ell(\hat{f}^{(i)}(c),y) \right]\nonumber
    \end{align}
    hence the tumor classifier recovers its performance \eqref{practical}. 
    
    We emphasize that our SST network does not require the dataset of institute $j$ to train both $\hat{f}^{(i)}$ and $\tau$. To make $\tau$ independent of institute $j\in I$, we employ the gray normalization $G:\mathcal{C}\to\mathcal{G}$ and train \emph{stain-style generator} $\zeta : \mathcal{G}\to\mathcal{C}$ such that $\tau = \zeta\circ G$, as illustrated in Figure \ref{fig:overview}.

    
    
    
    
    
    
    

    \subsection{Stain-Style Generator by Conditional GAN}

        \begin{figure}[t]
        \centering
        \includegraphics[width=1\textwidth]{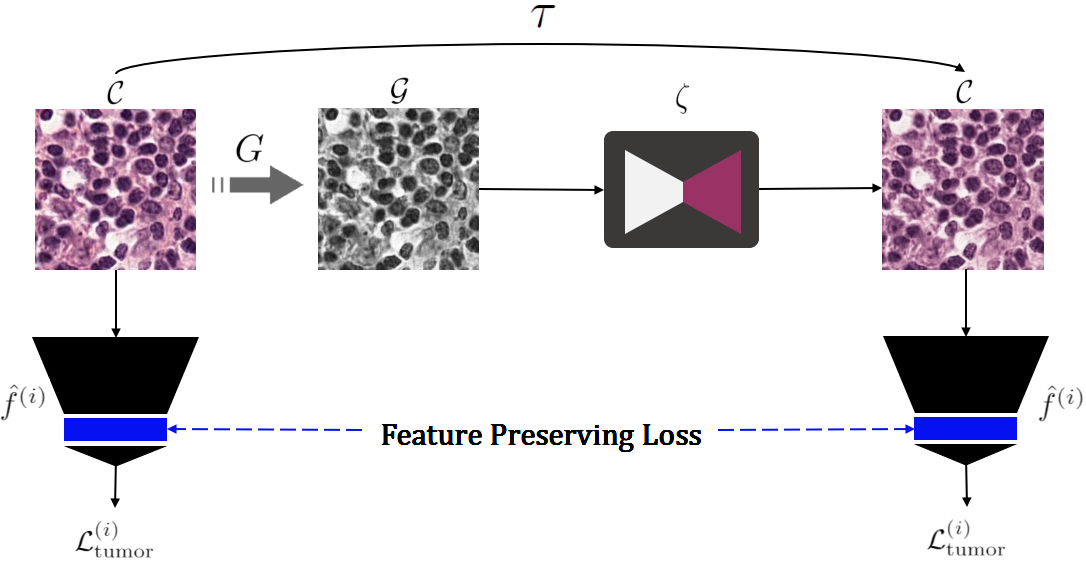}
        \caption{Illustration of feature-preserving loss. We use the global average pooled layers of input and generated image as the input of feature preserving loss.}
        \label{fig:fp_loss}
    \end{figure}
    To train the style-generator $\zeta$, we introduce three loss functions (a) reconstruction loss, (b) GAN loss, and (c) feature-preserving loss. 
    
    \subsubsection{Reconstruction Loss}

    Restricted to the initial $\phi^{(i)}$, SST network $\tau$ should be an reconstruction map i.e. $\tau\circ\phi^{(i)} = \phi^{(i)}$. Hence we apply a reconstruction loss to minimize the $L_{2}$-distance between $\tau\circ \phi^{(i)}$ and its original image $\phi^{(i)}$ using the architecture from \cite{quan2016fusionnet} which has very deep structure with short-cut and skip connections. The reconstruction loss is denoted by
    \begin{align*}
    \mathcal{L}_{\text{Recon}} (\tau, \phi^{(i)}) :=  \mathbb{E}\Vert \tau(\phi^{(i)}) - \phi^{(i)} \Vert_2
    \end{align*}
    
    \subsubsection{Conditional GAN Loss}
    
    As Pathak \emph{et al}. showed in \cite{pathak2016context}, mixing GAN loss \cite{goodfellow2014generative} with some traditional loss, such as $\mathcal{L}_{\text{recon}}$, improves the performance of generator. Since we have labeled images, conditional GAN \cite{mirza2014conditional} was applied instead of \cite{goodfellow2014generative}. By means of GAN, $\zeta$ is to learn a mapping from $\mathcal{G}$ to $\mathcal{C}$ and to trick the discriminator $\mathcal{D}$. Here $\mathcal{D}$ is to distinguish between fake and real images using the architecture from DCGAN \cite{radford2015unsupervised}. We use the following GAN loss
    
    \begin{align*}
    \mathcal{L}_{\text{GAN}} (\tau, \phi^{(i)}) := \mathbb{E}[\log \mathcal{D}(\zeta, \phi^{(i)})]+ \mathbb{E}[\log(1 - \mathcal{D}(\zeta, \phi^{(i)}))]
    \end{align*}
    While $\mathcal{D}$ learns to maximize $\mathcal{L}_{\text{GAN}}$, $\zeta$ tries to minimize it until both arrives at its optimal state.
    Through the above procedure, every stained image might be transferred to have the desired stain-style. 
    However, this approach often tend to make frequent color images independent of histological pattern. This phenomena is called mode collapse (of GANs) which possibly interrupt achieving \eqref{transfer}. Therefore we need an additional loss function. 
    
    \subsubsection{Feature-preserving Loss}\label{fp}
    
    As in \eqref{transfer}, in the optimal state $\tau^{*}$, an output of SST network $\tau^{*}\circ\phi^{(j)}$ should approximate target $\phi^{(i)}$. By the means of Kullback-Leibler divergence, \eqref{transfer} can be restated by
    \begin{align*}
    \tau^{*}=\arg \min_{\tau}\mathbb{KL}\left[\mathbb{P}^{(i)}(c|y) \Big\Vert \mathbb{P}^{(j)}(\tau^{-1}(c)|y)\right]
    \end{align*}
    To obtain $\tau^{*}$, having \eqref{approx} in mind, we employ the feature-presearving loss
    \begin{align*}
    \mathcal{L}_{\text{FP}} (\tau, \phi^{(i)}) := \mathbb{KL}\left[\mathcal{F}\left(\hat{f}^{(i)}(c)\right) \Big\Vert \mathcal{F}\left(\hat{f}^{(i)}(\tau(c))\right)\right]
    \end{align*}
    where $\mathcal{F}(\hat{f}^{(i)}(\cdot))$ indicates the feature of given color image extracted from the classifier $\hat{f}^{(i)}$. As illustrated in Figure \ref{fig:fp_loss}, the final layer before the activation function is used to examine feature vector, precisely, global average pooled layer. 
    
    Consequently, the overall loss function is
    \begin{align*}
    \mathcal{L} (\tau, \phi^{(i)}) := \lambda_{\text{Recon}}\mathcal{L}_{\text{Recon}}(\tau, \phi^{(i)}) + \mathcal{L}_{\text{GAN}}(\tau, \phi^{(i)}) + \lambda_{\text{FP}}\mathcal{L}_{\text{FP}}(\tau, \phi^{(i)})
    \end{align*}
    where $\lambda_{\text{Recon}}$, $\lambda_{\text{FP}}$ are the weights which are used to balance the update between different loss functions.

    \section{Experiment}

    
    We perform quantitative experiment in tumor classification to evaluate the SST network. To show the general performance of our method, we apply the extensions to vanilla models as well as conventional method. We have 4 baseline methods: \cite{reinhard2001color}, \cite{macenko2009method}, Histogram specification (HS) \cite{annadurai2007fundamentals} and WSI color standardization (WSICS) \cite{bejnordi2016stain}. 
    
    \subsection{Dataset}
    The Camelyon16 dataset is composed of 400 slides from two different institutes, Radbound and Utrecht. We use 180,000 patches for training, 20,000 for validation from Radbound and 140,000 patches for testing from Utrecht. The number of tumor and normal are the same. 
    Hypothesizing the training and validation dataset belong to a certain institue and the test set is from another one, we can merge every stain-style into the same space by applying the gray normalization. Both training and validation dataset are labeled, supervised learning can be applied to train the mapping from gray image to the colored one. 
    We used gray normalization based on \emph{Pillow} package of python which uses this formula $L = 0.299 \times R + 0.587 \times G + 0.144 \times B$.


    \subsection{Network Architecture}

    In this part, we explain each network structure of classifier network and stain-style generator which constitute SST network. 
    
    \subsubsection{Classifier Network}
    Classifier network carries out two tasks in experiment. Firstly, it is a discriminator which evaluates the performance of stain-style generator $\zeta$. Secondly, as already explained in subsection \ref{fp}, it works as a feature-extractor which is used in feature preserving loss $\mathcal{L}_{FP}$. We use ResNet-34 from \emph{torchvision} library in PyTorch as a framework. 
    
    \subsubsection{Stain-Style Generator}
     The generator network $\zeta$ is provided an image as an input instead of a noise vector. Therefore we can use FCN type architectures and U-Net is one of the most famous network among them. However, because of its limit of performance, we use FusionNet which has combined the advantages of U-Net and that of ResNet. Hyperparameters of network are set as same as \cite{quan2016fusionnet}. We adapt our discriminator architectures from \cite{radford2015unsupervised} which is based on VGG-Net without pooling layer. The hyperparameters of discriminator are the same as those in \cite{radford2015unsupervised}.

    \subsection{Result}
    
    \begin{figure}[t]
        \centering
        \includegraphics[width=1\textwidth]{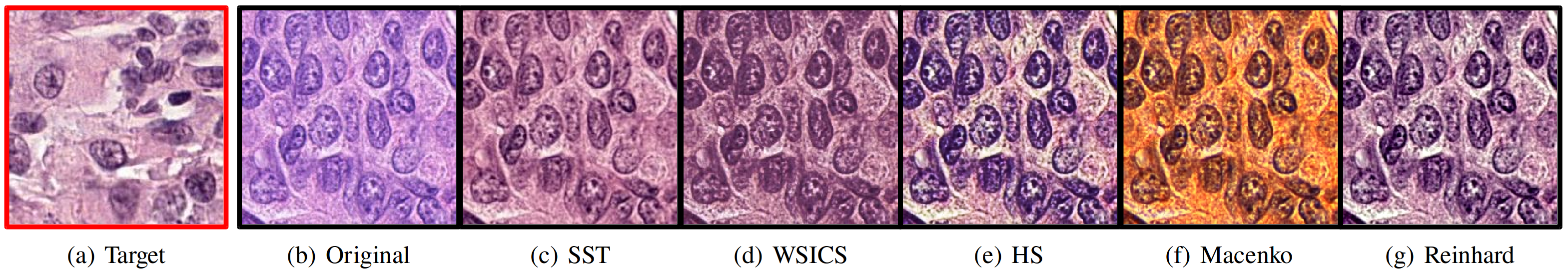}
        \caption{Comparison between SST and other stain normalization method: (a) Target image for transfer (b) Original input image to be transferred (c) SST (d) WSICS (e) HS (f) Marcenko (g) Reinhard }
        \label{fig:compare}
    \end{figure}

    Figure \ref{fig:compare} illustrates the result of each stain normalization method on a sample image. Target image comes from Radbound which is used for training the tumor classifier. Original image is sampled from Utrecht, used for testing the tumor classifier. Although there is no visual difference between outputs of each method, the classification performance on these color images varies significantly.
    Given the experiment results in Table \ref{table:confusion}, SST network successfully avoids the performance degradation. SST achieves the highest performance on original images on tumor classification with Area Under Curve(AUC) = 0.9185. This result shows that there are difference between visual judgment and the result of classifier. In case of WSICS's result, which is most visually similar to SST's, the AUC score is worse than that of SST by about 30\%. On the other hand, Macenko, which was visually the worst, performs better than other methods except for SST. Conventional methods consider only the physical  features  of  input  images and lose patterns which are key features for classifier's decision making process. In contrast, SST maintains those key features, input image's own patterns, and also consider the color distribution of target images as well as the contextual information of original images.

    \begin{table}[h]
        \centering
                \caption{Performance of tumor classifier network on different stain normalization methods. SST network shows significant improvement compared to direct application to original (untransferred image) and outperforms the others.}
        \begin{tabular}{ |p{1.5cm} ||p{1.4cm} |p{1.4cm} |p{1.4cm} |p{1.4cm} |p{1.4cm} |p{1.4cm} |p{1.4cm} |}
            \hline
            Model & Target & Original & \textbf{SST} & WSICS & HS & Macenko & Reinhard \\  
            \hline\hline
            AUC& 0.9760& 0.8900& \textbf{0.9185} & 0.6408 & 0.4245 & 0.7169 & 0.5611\\ 
            \hline
            Precision& 0.9114& 0.8098& \textbf{0.8440} & 0.5989 & 0.4987 & 0.6983 & 0.6114\\ 
            \hline
            Recall& 0.9126& 0.8111& \textbf{0.8460} & 0.5957 & 0.4986 & 0.6956 & 0.6119\\ 
            \hline
            Specificity& 0.9583& 0.8014& \textbf{0.8371} & 0.6010 & 0.4162 & 0.6500 & 0.5471\\ 
            \hline
        \end{tabular}
        \label{table:confusion}
    \end{table}

    \section{Conclusion}
    
    In this work, we have presented a stain style transfer approach to stain normalization for histopathological images. To that end, we replace the stain normalization models with a generative model which learns certain stain-style distribution of training dataset. This stain style transfer network is considerably simpler than contemporaneous work, and produces more realistic results without any additional labeling or annotation for training as well as prior knowledge. Further, unlike conventional stain normalization, which acts independently of the tumor classifier, the proposed feature-preserving loss induces our coloration in a direction that does not affect the tumor classifier. We demonstate that our model is optimized for the performance of the tumor classifier and allows successful stain-style transfer.
    
    The style of chemical cell staining is mainly affected by structural information and morphology of cells rather than factors such as cell brightness. Based on these observation points, we converted the test image into a gray image and performed a stain style transfer process. While this method has the advantage of making the process simpler, it has also lost some information. To resolve the limitation, further investigation will assess direct stain style transfer approach from color image to color image. In addition, we hope to more closely examine parameters of our deep learning approach. Further, we will perform more rounds of hard negative mining and consider the reliability and reproducibility of the deep CNN models.

            
    \bibliographystyle{ieeetr}
    \bibliography{biblist}

\end{document}